\documentclass{article}

\usepackage{PRIMEarxiv}

\usepackage[utf8]{inputenc} % allow utf-8 input
\usepackage[T1]{fontenc}    % use 8-bit T1 fonts
\usepackage{hyperref}       % hyperlinks
\usepackage{url}            % simple URL typesetting
\usepackage{booktabs}       % professional-quality tables
\usepackage{amsfonts}       % blackboard math symbols
\usepackage{nicefrac}       % compact symbols for 1/2, etc.
\usepackage{microtype}      % microtypography
\usepackage{lipsum}
\usepackage{fancyhdr}       % header
\usepackage{graphicx}       % graphics
\usepackage{amssymb}
\usepackage{lineno}
\usepackage{amsmath}
\usepackage{bm}
\usepackage{algorithm}
\usepackage[american]{babel}
\usepackage{algorithmic}
\usepackage{subfig}
\usepackage{listings}
\graphicspath{{media/}}     % organize your images and other figures under media/ folder

\title{\emph{ADS}: Approximate Densest Subgraph for Novel Image Discovery
\thanks{\textit{\underline{Citation}}: 
\textbf{Authors. Title. Pages.... DOI:000000/11111.}} 
}

\author{
  Shanfeng Hu \\
  Northumbria University \\
  Newcastle upon Tyne, UK\\
  \texttt{shanfeng2.hu@northumbria.ac.uk} \\
}

\begin{document}
\maketitle

\begin{abstract}
The volume of image repositories continues to grow. Despite the availability of content-based addressing, we still lack a lightweight tool that allows us to discover images of distinct characteristics from a large collection. In this paper, we propose a fast and training-free algorithm for novel image discovery. The key of our algorithm is formulating a collection of images as a perceptual distance-weighted graph, within which our task is to locate the $K$-densest subgraph that corresponds to a subset of the most unique images. While solving this problem is not just NP-hard but also requires a full computation of the potentially huge distance matrix, we propose to relax it into a $K$-sparse eigenvector problem that we can efficiently solve using stochastic gradient descent (SGD) without explicitly computing the distance matrix. {We compare our algorithm against state-of-the-arts on both synthetic and real datasets, showing that it is considerably faster to run with a smaller memory footprint while able to mine novel images more accurately.}
\end{abstract}

% keywords can be removed

\section{Introduction}
\label{sec:introduction}
We are naturally attracted by visually novel scenes in our daily lives \cite{yang2015learning,clark2018nice}. While various photo management softwares can provide the functionalities of chronologically and semantically indexing these visual assets for easy access, they still lack the basic ability to let a user navigate some of the most memorable images she has taken before. This feature may not just satisfy our own curiosity of recalling "old surprises" but can also prompt sufficiently novel images to be shared on social media for views attraction. Posting less interesting images, in contrast, can lead to lower user engagement on social media \cite{li2020picture}. Despite the availability of various content-based retrieval of images using state-of-the-art deep learning technologies \cite{zhang2021deep,zhang2021deformable}, we still lack a lightweight and efficient tool to automatically mine images of distinct nature.

\textbf{Related Work}. While assessing whether an image is of great interest or not is subject to personal perception to some extend, there are work on image memorability showing that the judgements of human participants are actually quite consistent among each other \cite{isola2011makes,jaegle2019population}. This has since then sparked a range of efforts that attempt to predict the memorability of images using machine learning and deep learning techniques \cite{isola2011makes,kim2013relative,khosla2015understanding,jing2018framework,squalli2018deep,akagunduz2019defining,perera2019image,basavaraju2020image}. Despite the success, these learning-based methods can suffer a few limitations. First, they require a ground-truth set of human-annotated memorability scores for training, which can be costly to collect. Second, the predictive power of the trained models depends on whether test images are sufficiently similar to those in the training set. As new images arrive, distributional shifts can happen and as a result incur prediction errors. And third, the existing methods make predictions on an image-by-image case, without comparing the relative novelty of images as a whole. However, this problem would be more natural to solve as a ranking rather than a regression task.

Visual saliency detection is another area of research that could help identify images that significantly deviate from others within a collection. However, most saliency detection methods only look for unique regions or objects within individual images \cite{itti1998model,hou2007saliency,yan2018unsupervised,wang2021salient,lv2022semi}, which lack the objective of assessing the uniqueness of whole images. While a variety of group-level saliency detection approaches have been proposed \cite{zhang2015co,zhang2018review,wang2019robust}, they are mainly designed to search for co-occurring prominent objects of an image collection. Salient features have also been used for image memorability prediction \cite{mancas2013memorability,khosla2015understanding}, but it remains unclear how to formalize the relationship between local saliency and global ranking of image novelties.

\textbf{Our Contributions}. In view of the aforementioned challenges, we propose a lightweight algorithm for novel image discovery that is mathematically clear to understand, practically simple to implement, fast to run, and free of training (hence requiring no human annotated labels). Different from the existing work on image memorability prediction, we formulate the problem as searching for a global ranking of images in a collection that maximizes a total novelty score. The key idea of our algorithm is representing a collection of images as an undirected complete graph, with the nodes being the images and each edge weighted using the perceptual distance of the two corresponding images. This allows us to identify a subset of the $K$ most distinct images by finding the $K$-densest subgraph within the complete graph \cite{feige2001dense}. The total novelty score is the sum of the mutual distances among selected images.

Due to the NP-hardness of the $K$-densest subgraph problem \cite{feige2001dense}, a wide range of approximate algorithms have been proposed to solve it efficiently \cite{tsourakakis2013denser,tsourakakis2015k,jazayeri2021frequent}. However, a main limitation of these methods is that they require a full graph representation of data, which in our case would mean that we need to pre-compute the pairwise distance matrix of all images in a collection for novelty finding. If we have 10,000 images in the collection, then we will have a $10000 \times 10000$ dense matrix that is trivially intractable to handle on a resources-limited device such as our phone. To address this challenge, we propose to relax the original problem into a half-discrete, half-continuous version, which looks for a $K$-sparse eigenvector that maximizes a weighted sum of the mutual distances among the selected images. The higher a weight (an element in the eigenvector), the more novel the corresponding image. A zero weight indicates that the image is not selected.

The proposed relaxation allows us to solve the problem much more efficiently using SGD, without explicitly computing or storing the full distance matrix of images. The key step is that we can approximate the full gradient evaluation at each step with a Monte-Carlo estimate, which only requires sampling a mini-batch of images for comparison with each image to update its weight. While SGD and its variants are de facto optimizers for deep learning \cite{bottou2010large}, we further equip it with a clipping operator that prunes unpromising images early on while ensuring the $K$-sparsity constraint of the solved eigenvector. Due to the scalability of SGD, we are able to efficiently discover a required amount of novel images from a potentially large image repository.

\section{Novel Image Discovery}
\label{sec:novel-image}

\subsection{Problem Formulation}
\label{subsec:problem}
Suppose we are given a set of $N$ images, $\mathcal{C} = \{\bm{x}_i\}_{i=1}^{N}$. For each image $\bm{x}_i$ we are able to extract a $M$-dimensional feature vector, $\phi(\bm{x}_i) \in \mathbb{R}^M$, which represents the visual and semantic contents of the image. The problem of novel image discovery can then be formulated as finding a subset of at most $K$ images $\mathcal{S}(\mathcal{C})=\{\bm{x}_{(k)}\}_{k=1}^{K}$ within the collection $\mathcal{C}$ that are the most distinct from each other as well as from the remaining ones in the feature space. As explained in Section \ref{sec:introduction}, the existing work on image memorability prediction treats this problem as an individual-based regression task \cite{isola2011makes,jing2018framework,squalli2018deep,akagunduz2019defining,perera2019image}, without optimizing the global ranking of images within a collection. Instead, we formalize it as a mathematical optimization problem:
\begin{align}
	\label{eqn:discrete-problem}
	\mathcal{S}(\mathcal{C}) = \underbrace{\underset{\mathcal{S}}{\text{arg max\ }} J(\mathcal{S}) \text{\ s.t. $\mathcal{S} \subset \mathcal{C} \land |\mathcal{S}| \le K$}}_{\text{novel image discovery problem}}
\end{align}
with $|\mathcal{S}|$ being the cardinality of the solution set and the objective function $J(\mathcal{S})$ is defined as
\begin{equation}
	\label{eqn:discrete-objective}
	J(\mathcal{S}) = \underbrace{\sum_{\bm{x}_i \in \mathcal{S}}{\sum_{\bm{x}_j \in \mathcal{S}}{d(\bm{x}_i, \bm{x}_j)}}}_{\text{total novelty score}} 
\end{equation}
in which we use the Euclidean distance in the feature space to measure the closeness for every pair of images
\begin{equation}
	\label{eqn:distance}
	d(\bm{x}_i,\bm{x}_j) = \|\phi(\bm{x}_i)-\phi(\bm{x}_j)\|
\end{equation}
We refer to the objective (\ref{eqn:discrete-objective}) as the \emph{total novelty score} as it measures the sum of the mutual feature distances among all selected images. The higher the score, the more separated the images in $\mathcal{S}$ from each other. We are searching for the $K$ most unique images within $\mathcal{C}$ to maximize the score.

If we view the given collection of images $\mathcal{C}$ as an undirected complete graph in which each image is a node and the edge between every pair of images is weighted using their feature distance (\ref{eqn:distance}), then the problem (\ref{eqn:discrete-problem}) amounts to solving for the $K$-densest subgraph of the complete graph \cite{feige2001dense}. This problem is well known for its NP-hardness, for which various approximate solutions have been proposed in the past \cite{tsourakakis2013denser,qin2015locally, tsourakakis2015k, chekuri2022densest}. However, these methods generally require that every pair of data points need to be compared in the solution process, which may not be feasible when the underlying set $\mathcal{C}$ is very large (e.g., containing over 10,000 images). This presents a particular computational hurdle on resources-limited devices such as our phone. We address this challenge in Section \ref{subsec:algorithm} using our proposed algorithm.

\subsection{The Proposed Algorithm}
\label{subsec:algorithm}
Here, we present a simple and scalable algorithm for approximately solving problem (\ref{eqn:discrete-problem}) to find the subset of most unique images within a collection. We start with relaxing the problem into a half-discrete, half-continuous version in Section \ref{subsubsec:relaxation}, which we then efficiently solve using SGD and sparsity clipping in Section \ref{subsubsec:stochastic}. We analyze the computational complexity of our algorithm in Section \ref{subsec:complexity}. We give the pseudocode in Algorithm (\ref{alg:ads}).

\begin{algorithm}
	\caption{Approximate Densest Subgraph (ADS)}
	\label{alg:ads}
	\begin{flushleft}
		\textbf{Input}: $\mathcal{C} = \{\bm{x}_i\}_{i=1}^{N}$, a collection of $N$ images. \\
		
		\textbf{Parameters}: $K$, number of novel images to be found from $\mathcal{C}$; $\phi$, feature extractor; $E$, epoch size (default: 20); $J$, batch size (default: 16); $\alpha$, learning rate (default: 0.001); $\beta$, momentum (default: 0.9). \\
		
		\textbf{Output}: $\bm{s}(\mathcal{C})$, a $K$-sparse vector of novelty weights.
	\end{flushleft}
	
	\begin{algorithmic}[1] %[1] enables line numbers
		\STATE $\{\phi(\bm{x}_i)\}_{i=1}^{N} \xleftarrow{\phi} \mathcal{C}$
		
		\STATE Let $t=1$, $\bm{s}^0=\frac{1}{\sqrt{N}}[1,1,\cdots,1]^T$, $\Delta^0_i =  0$ for $1 \le i \le N$
		
		\WHILE{$t \le E$}
		\STATE Let $i=1$
		\WHILE{$i \le N$}
		\STATE $\mathcal{J}^t_i \leftarrow \text{random\_choice}(\{i|\bm{s}_i^{t-1} > 0\}, J)$
		
		\STATE Let $\Delta_i^t = 0$
		\FOR{$j \in \mathcal{J}^t_i$}
		\STATE $\Delta_i^t \leftarrow \Delta_i^t + \|\phi(\bm{x}_i)-\phi(\bm{x}_j)\|\bm{s}^{t-1}_j$
		\ENDFOR
		
		\IF{$t \ge 2$}
		\STATE $\Delta_i^t \leftarrow (1-\beta)\Delta_i^t + \beta \Delta_i^{t-1}$
		\ENDIF
		
		\STATE $\bm{s}^t_i \leftarrow \bm{s}^{t-1}_i + 2\alpha \frac{N}{J}\Delta_i^t$
		\STATE $\Delta_i^{t-1} \leftarrow \Delta_i^t$
		
		\STATE $i \leftarrow i+1$
		\ENDWHILE
		
		\STATE Let $K^t = \lfloor N - \frac{N-K}{E}t\rfloor$
		\STATE $\bm{s}^t \leftarrow \bm{s}^t \circ [I(1 \in \text{top-}K^t(\bm{s}^t)), \cdots, I(N \in \text{top-}K^t(\bm{s}^t))]^T$
		
		\STATE $\bm{s}^t \leftarrow \frac{\bm{s}^t}{\|\bm{s}^t\|}$
		
		\STATE $t \leftarrow t+1$
		\ENDWHILE
		\STATE $\bm{s}(\mathcal{C}) \leftarrow \bm{s}^T$
		\STATE \text{Return $\bm{s}(\mathcal{C})$}
	\end{algorithmic}
\end{algorithm}

\subsubsection{\textbf{Sparse Continuous Relaxation}}
\label{subsubsec:relaxation}
Let us assume for the moment that we can afford to compute and store the full distance matrix of images in the collection $\mathcal{C}$, $D \in \mathbb{R}_{\ge 0}^{N \times N}$, with each element $D_{ij}=d(\phi(\bm{x}_i),\phi(\bm{x}_j))$ as computed in equation (\ref{eqn:distance}).  It is easy to see that problem (\ref{eqn:discrete-problem}) can be equivalently expressed as the following one:
\begin{equation}
	\label{eqn:discrete-problem-matrix}
	\bm{s}(\mathcal{C}) = \underbrace{\underset{\bm{s}}{\text{arg max\ }} \bm{s}^TD \bm{s} \text{\ s.t. $\bm{s} \in \{0,1\}^N \land |\bm{s}|_0 \le K$}}_{\text{reformulation of (\ref{eqn:discrete-problem}) using matrix form}}
\end{equation}
in which the optimization variable $\bm{s}$ is restricted to be a vector of 0 and 1 of length $N$, with $\bm{s}_i=1$ indicating $\bm{x}_i \in \mathcal{S}(\mathcal{C})$ and $\bm{s}_i=0$ otherwise. The objective $J(\bm{s})=\bm{s}^TD\bm{s}=\sum_{i=1}^{N}{\sum_{j=1}^{N}{D_{ij}\bm{s}_i\bm{s}_j}}$ hence equals to the original one in (\ref{eqn:discrete-objective}). The L0 norm of the vector, $|\bm{s}|_0$, namely the number of non-zero elements in the vector, is required to be no greater than the maximal number of selected images $K$.

While the new formulation (\ref{eqn:discrete-problem-matrix}) is still NP-hard to solve, we now have the opportunity to relax it into a half-discrete, half-continuous version by allowing each element of the solution vector to take on non-negative real values:
\begin{equation}
	\label{eqn:relaxation}
	\bm{s}(\mathcal{C}) = \underbrace{\underset{\bm{s}}{\text{arg max\ }} \bm{s}^TD \bm{s} \text{\ s.t. $\bm{s} \in \mathbb{R}_{\ge 0}^N \land \|\bm{s}\|=1 \land |\bm{s}|_0 \le K$}}_{\text{sparse continuous relaxation of (\ref{eqn:discrete-problem-matrix})}}
\end{equation}
In this case, $\bm{s}_i > 0$ would mean that the $i$-th image is selected into the subset $\mathcal{S}(\mathcal{C})$ with a certain weight while $\bm{s}_i = 0$ indicates that the image is not selected. Since $K$ is normally chosen to be much smaller than $N$, the solution is therefore enforced to be sparse while maximizing the weighted sum of the mutual feature distances among the selected images. Intuitively speaking, the more novel an image $\bm{x}_i$ given that it is selected, the higher its score $\bm{s}_i(\mathcal{C})$, providing a fine-grained ranking of images within $\mathcal{S}(\mathcal{C})$ according to their relative uniqueness. Solving the original problem (\ref{eqn:discrete-problem}) or (\ref{eqn:discrete-problem-matrix}), however, does not enjoy this feature as the solution would be binary with no ranking information generated.

It is interesting to note that when $K=N$ (i.e., the sparsity constraint is dropped), the relaxed problem (\ref{eqn:relaxation}) is further reduced to solving for the unit-length principal eigenvector of the distance matrix $D$ that is associated with the largest eigenvalue. The Perron–Frobenius theorem \cite{pillai2005perron} states that due to the symmetry and non-negativity of $D$, the solution is guaranteed to be non-negative hence automatically satisfying the  constraint $\bm{s} \in \mathbb{R}_{\ge 0}^N$. The principal eigenvector can be found through the classical power iteration: $\bm{s}^t = \frac{D\bm{s}^{t-1}}{\|D\bm{s}^{t-1}\|}$, for which $\bm{s}^0$ can be initialized to be the unit-length uniform vector and the iteration can be stopped at step $t$ when $\|\bm{s}^t-\bm{s}^{t-1}\| < \epsilon$, with $\epsilon$ being a small constant for convergence checking. However, this requires the $N \times N$ dense distance matrix to be fully instantiated for iteration, which can be costly when $N$ is large. Also, the method cannot guarantee the $K$-sparsity of the sought solution. 

\subsubsection{\textbf{Stochastic Approximation}}
\label{subsubsec:stochastic}
Here, we address the aforementioned challenges of power iteration (Section \ref{subsubsec:relaxation}) with the following steps. First, we notice that by temporarily dropping the constraints in problem (\ref{eqn:relaxation}), we can locally maximize the objective by following the gradient direction:
\begin{align}
	\bm{s}^t &= \bm{s}^{t-1} + \alpha \frac{\partial (\bm{s}^TD\bm{s})}{\partial \bm{s}}|_{\bm{s}^{t-1}} \label{eqn:gradient-1}\\
	&= \bm{s}^{t-1} + 2\alpha D\bm{s}^{t-1} \label{eqn:gradient-2}
\end{align}
where $\alpha > 0$ is the learning rate (or step size). When $\alpha$ is sufficiently large, the update rule (\ref{eqn:gradient-2}) has a roughly similar effect of running one step of power iteration. While the gradient evaluation at each step is exact, it involves the full distance matrix and can be computationally heavy. We choose to approximate it using an Monte-Carlo estimate of the gradient for each component of the solution vector:
\begin{align}
	\bm{s}^t_i &= \bm{s}^{t-1}_i + 2\alpha \sum_{j=1}^{N}{D_{ij}\bm{s}^{t-1}_j} \label{eqn:component-1} \\
	&\approx \underbrace{\bm{s}^{t-1}_i + 2\alpha \left[(1-\beta) \frac{N}{|\mathcal{J}_i^t|} \sum_{j \in \mathcal{J}_i^t}{D_{ij}\bm{s}^{t-1}_j} + \beta\Delta^{t-1}_i\right]}_{\text{stochastic gradient descent with momentum}} \label{eqn:component-2}
\end{align}
where the step from (\ref{eqn:component-1}) to (\ref{eqn:component-2}) is done by replacing the costly exact sum with the rescaled sum over a small mini-batch of randomly picked images $\mathcal{J}_i^t \subset \{1,2,\cdots,N\}$ for image $\bm{x}_i$ at step $t$. We also maintain an exponential moving average of the historical sums $\Delta^{t-1}_i$ for each image $\bm{x}_i$ to stabilize the SGD learning process. $\beta$ controls the strength of the momentum.

The intuition of (\ref{eqn:component-2}) is that at each step $t$ when we update the $i$-th image's novelty weight $\bm{s}^{t-1}_i$, we randomly draw a small sample of images from the collection $\mathcal{C}$ and compare each of them with $\bm{x}_i$. If an image $\bm{x}_j$ for $j \in \mathcal{J}_i^t$ has a large distance from  $\bm{x}_i$ and its weight $\bm{s}^{t-1}_j$ is also high, then we increase $\bm{s}^{t-1}_i$ by a large amount in the update because the image $\bm{x}_i$ is found to be distinct from $\bm{x}_j$ that is also likely to be novel. Otherwise, when either the two images are very close in the feature space or if $\bm{x}_j$ is not sufficiently distinct from the  previous step, we only increase the weight by a comparatively small size to indicate that $\bm{x}_i$ may not be a good candidate of novelty given the evidence of the current step.

To impose the $K$-sparsity constraint, we apply the following clipping operation after executing each step of (\ref{eqn:component-2}):
\begin{equation}
	\label{eqn:clipping}
	\bm{s}^t_i =\begin{cases}
		\bm{s}^t_i \text{\ if $i \in \text{top-}K^t(\bm{s}^t)$} \\
		0 \text{\ otherwise}
	\end{cases}
\end{equation}
where $\text{top-}k^t(\cdot)$ returns the indices of the $K^t$ largest elements of a given vector. After truncating the remaining least weights of $\bm{s}^t$ to zeros, we normalize it to become a unit-length vector for the next iteration: $\bm{s}^t \leftarrow \frac{\bm{s}^t}{\|\bm{s}^t\|}$. It can be seen that by gradually decreasing $K^t$ from $N$ down to the required number of selected images $K$ during iteration, we can effectively filter out images that are not sufficiently novel in the SGD learning process. This also suggests that when we sample $\mathcal{J}_i^t$ for image $\bm{x}_i$ at step $t$, we only need to draw those images with a non-zero weight (i.e., $\bm{s}_j^{t-1} > 0$) as those with a zero weight do not contribute to the update (\ref{eqn:component-2}).

Our algorithm as described in (\ref{alg:ads}) has a few meta parameters to specify for each problem instance. The foremost one is $K$ - the number of novel images to be found and returned. A feature extractor $\phi(\bm{x})$ is also needed to produce dense embeddings of images for perceptual distance calculation. The SGD learning rate $\alpha$ controls the step size of each update, the momentum parameter $\beta$ stabilizes the learning process, and $J=|\mathcal{J}_i^t|$ is the batch size that determines the number of random samples for gradient estimation. Following the practice of deep learning, we also introduce the epoch size $E$ as the maximal number of learning steps to be executed. Given a fixed $E$, we can schedule the sparsity $K^t$ to gradually decrease from $N$ to $K$ through the learning process. One simple heuristic is $K^t = \lfloor N - \frac{N-K}{E}t\rfloor$ for $1 \le t \le E$.

\subsection{Complexity Analysis}
\label{subsec:complexity}
It is clear from the pseudocode (\ref{alg:ads}) that our algorithm does not need to store the $N \times N$ distance matrix for novel image discovery. When $N$ is large (e.g., over 10,000), the run-time space saving is huge compared with any algorithms that require a full realization of the matrix (or the underlying graph structure) to work. The time complexity of our algorithm is $\Theta(E(NJM+N\log N))$, where $E$ is the epoch size, $J$ is the batch size, and $M$ is the feature dimension. To sense check this, the time complexity of the classical power iteration is $\Theta((\frac{M}{2}+L)N^2)$, where $L$ is the required number of steps towards convergence. It can be seen that our algorithm scales well with the growth of $N$.

\section{Results and Discussion}
\label{sec:results}
We put our algorithm to the test of synthetic datasets in Section \ref{subsec:synthetic} and demonstrate its novelty detection accuracy by identifying images of novel classes in Section \ref{subsec:novel-class}{, while analysing its gradient approximation behaviour in these cases.} {We also compare it against the state-of-the-art methods on increasingly larger datasets in Section \ref{subsec:comparison}.} Without mentioned otherwise, we run all experiments using the default parameters as stated in the pseudocode (\ref{alg:ads}). 

\begin{figure*}
	\centering
	\subfloat[a][After 1 SGD step]{\includegraphics[scale=0.3]{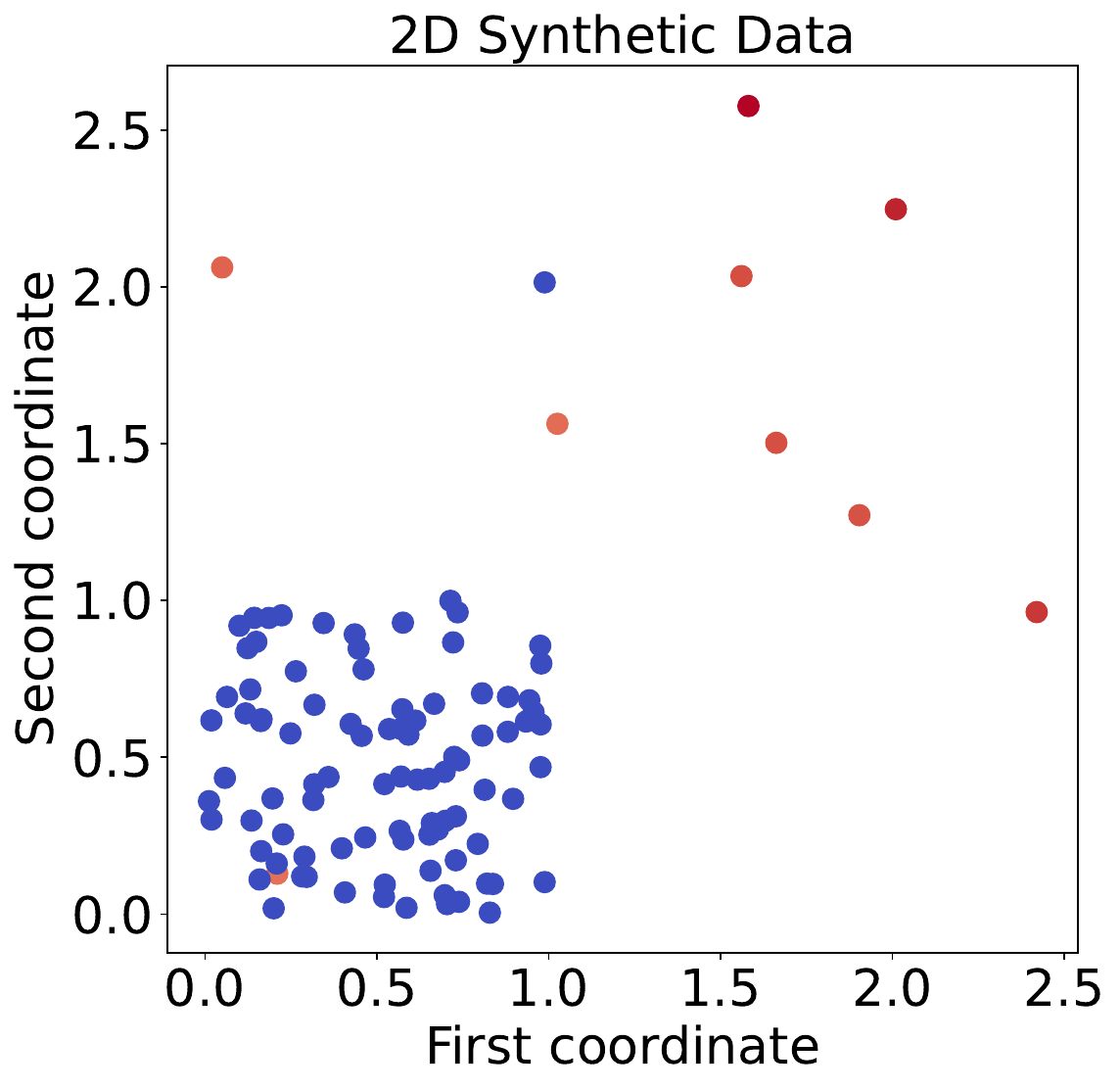} \label{fig:synthetic-data-1-1}}
	\subfloat[b][After 2 SGD steps]{\includegraphics[scale=0.3]{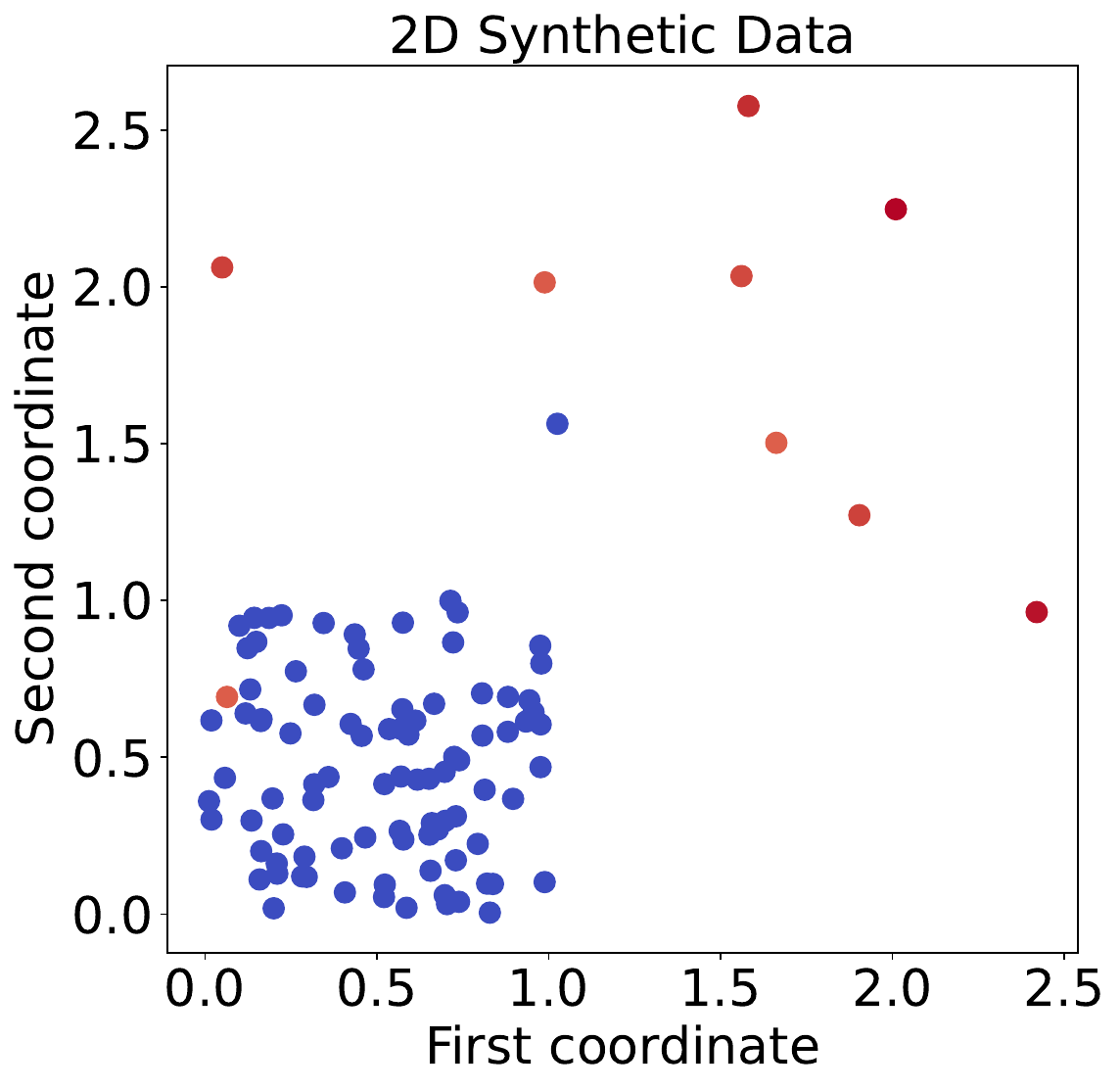} \label{fig:synthetic-data-1-2}}
	
	\subfloat[c][After 3 SGD steps]{\includegraphics[scale=0.3]{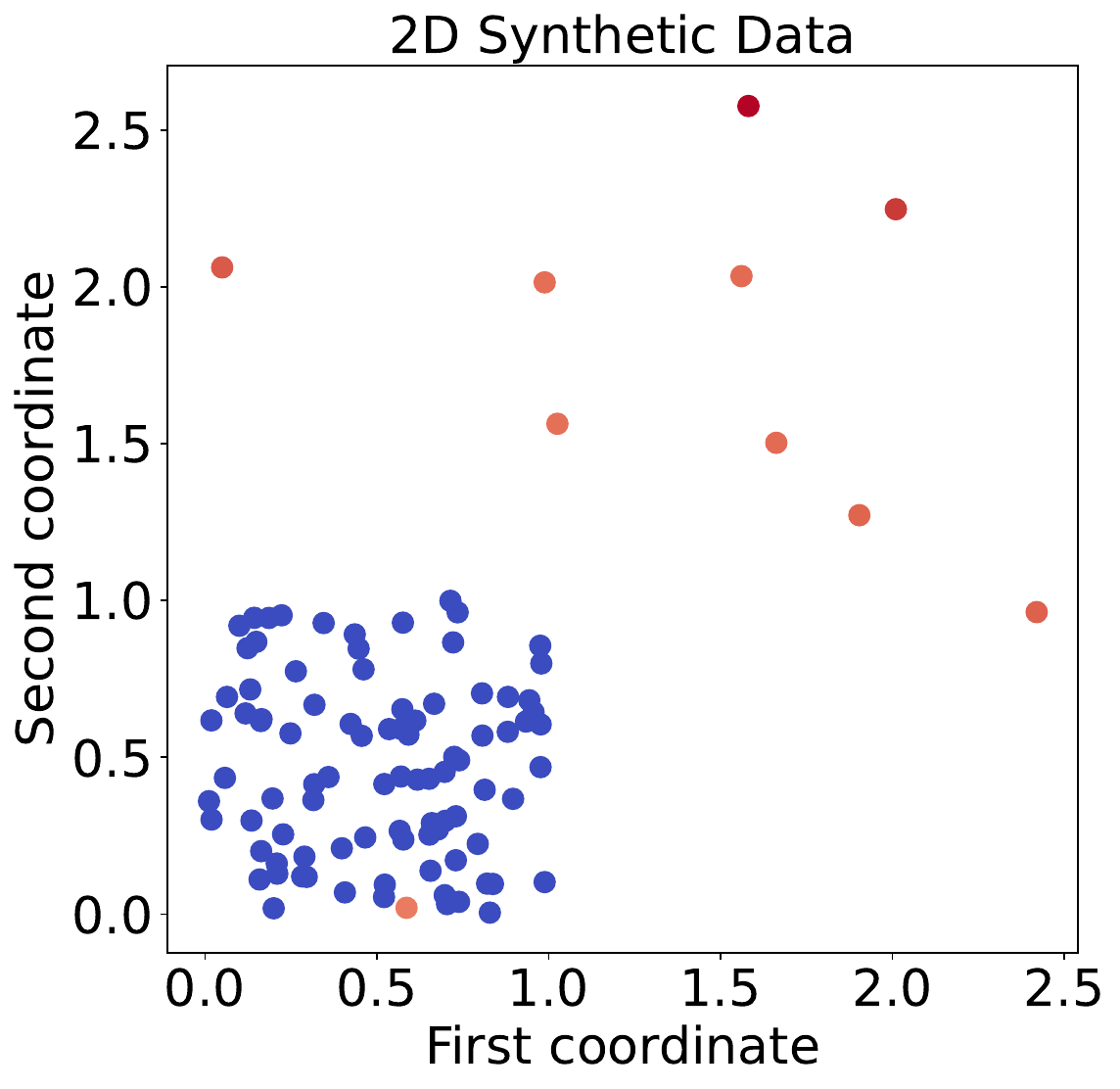} \label{fig:synthetic-data-1-3}}
	\subfloat[d][After 4 SGD steps]{\includegraphics[scale=0.3]{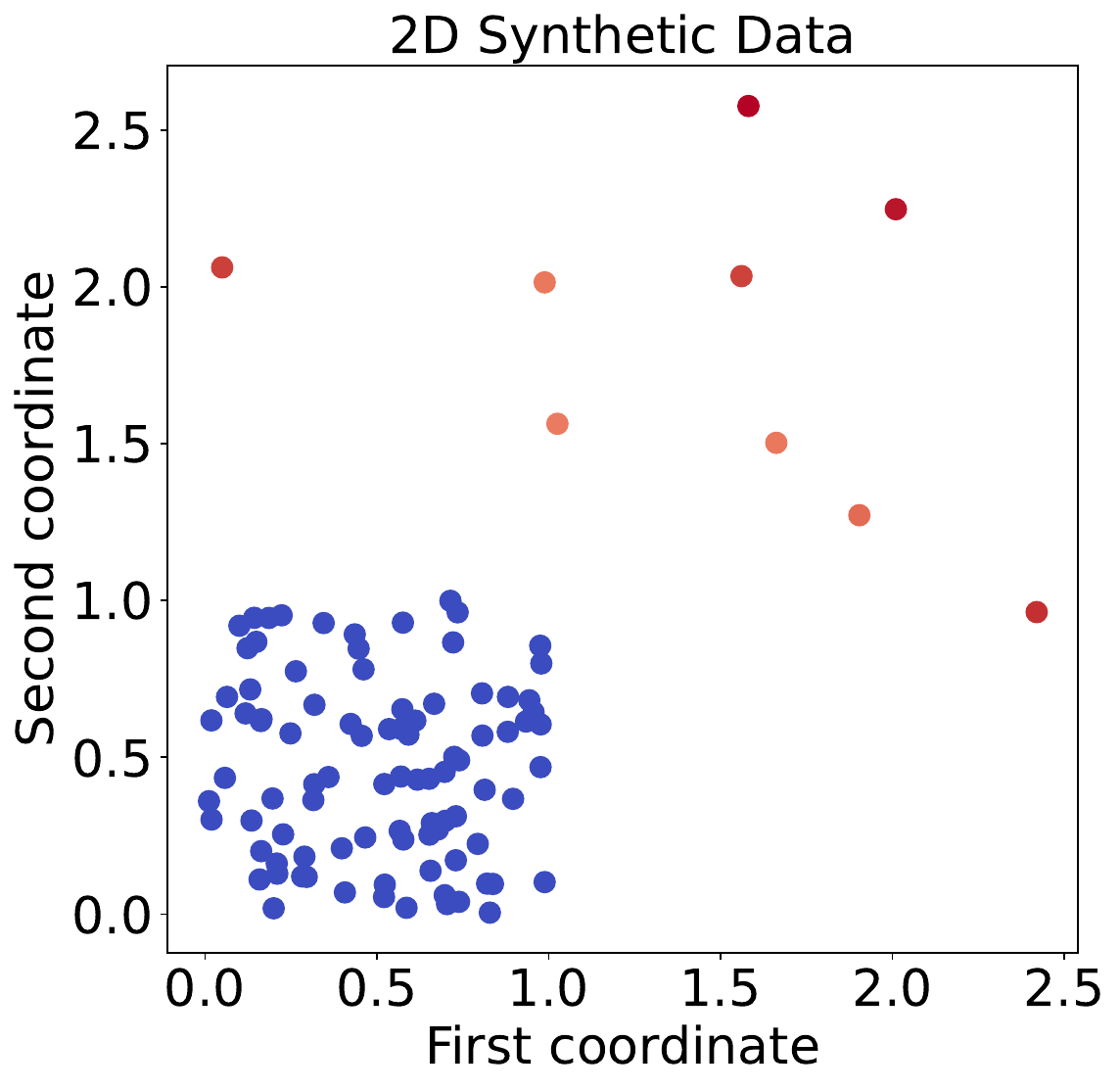} \label{fig:synthetic-data-1-4}}
	\caption{A 2-dimensional synthetic dataset. The dataset contains 100 uniformly generated 2D points with a few of them perturbed to become anomalies. The colour of each data point is from cold to warm indicating the corresponding value of the sparse eigenvector computed by our algorithm (\ref{alg:ads}). The sparsity parameter $K$ is set to 9 (the exact number of anomaly points). The epoch size $E$ is displayed under each sub figure from (a) to (h). The remaining parameters are set to their default values.} \label{fig:synthetic-data-1}
\end{figure*}

\begin{figure*}
	\centering
	\subfloat[e][After 1 SGD step]{\includegraphics[scale=0.3]{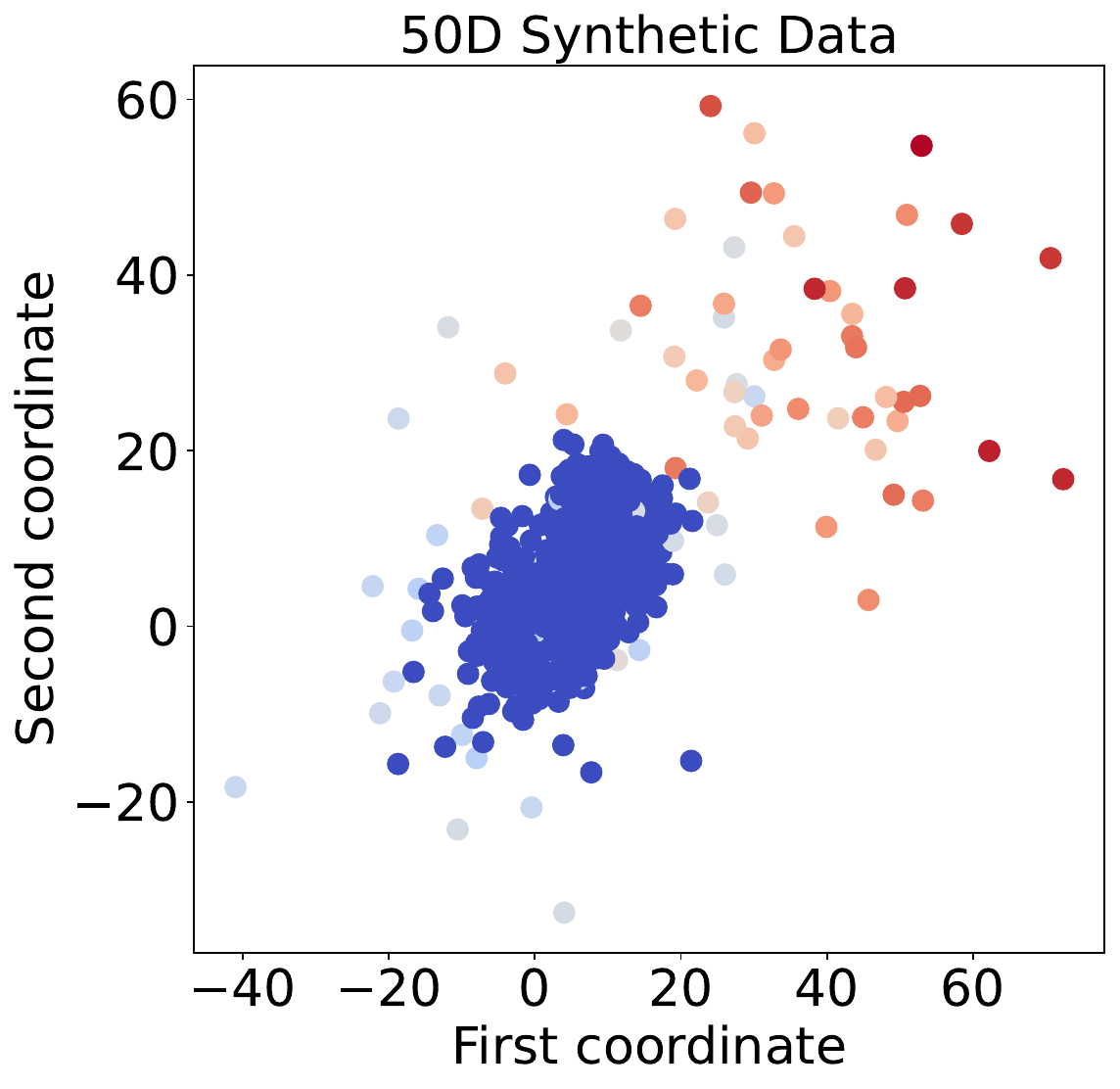} \label{fig:synthetic-data-2-1}}
	\subfloat[f][After 5 SGD steps]{\includegraphics[scale=0.3]{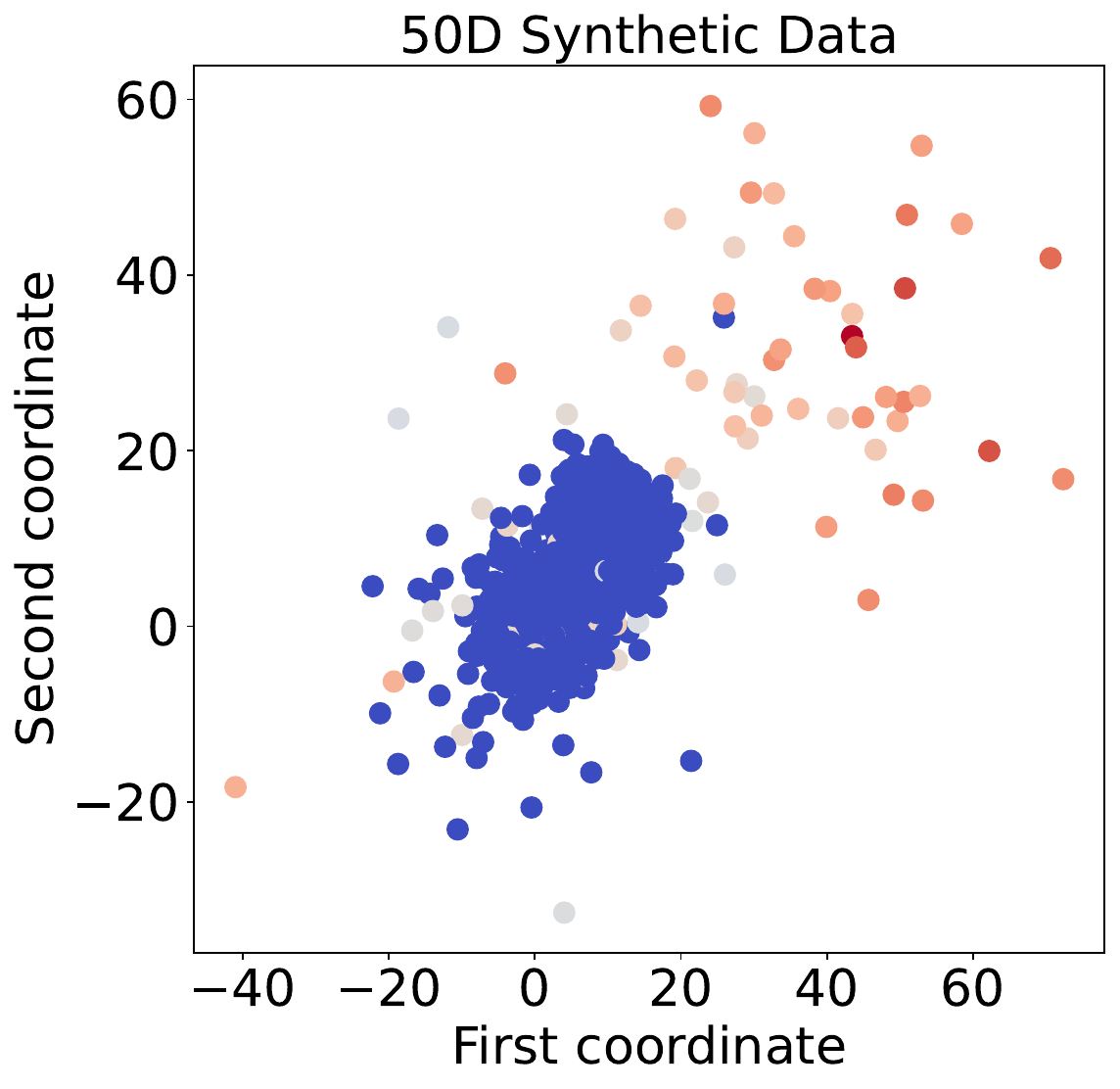} \label{fig:synthetic-data-2-5}}
	
	\subfloat[g][After 10 SGD steps]{\includegraphics[scale=0.3]{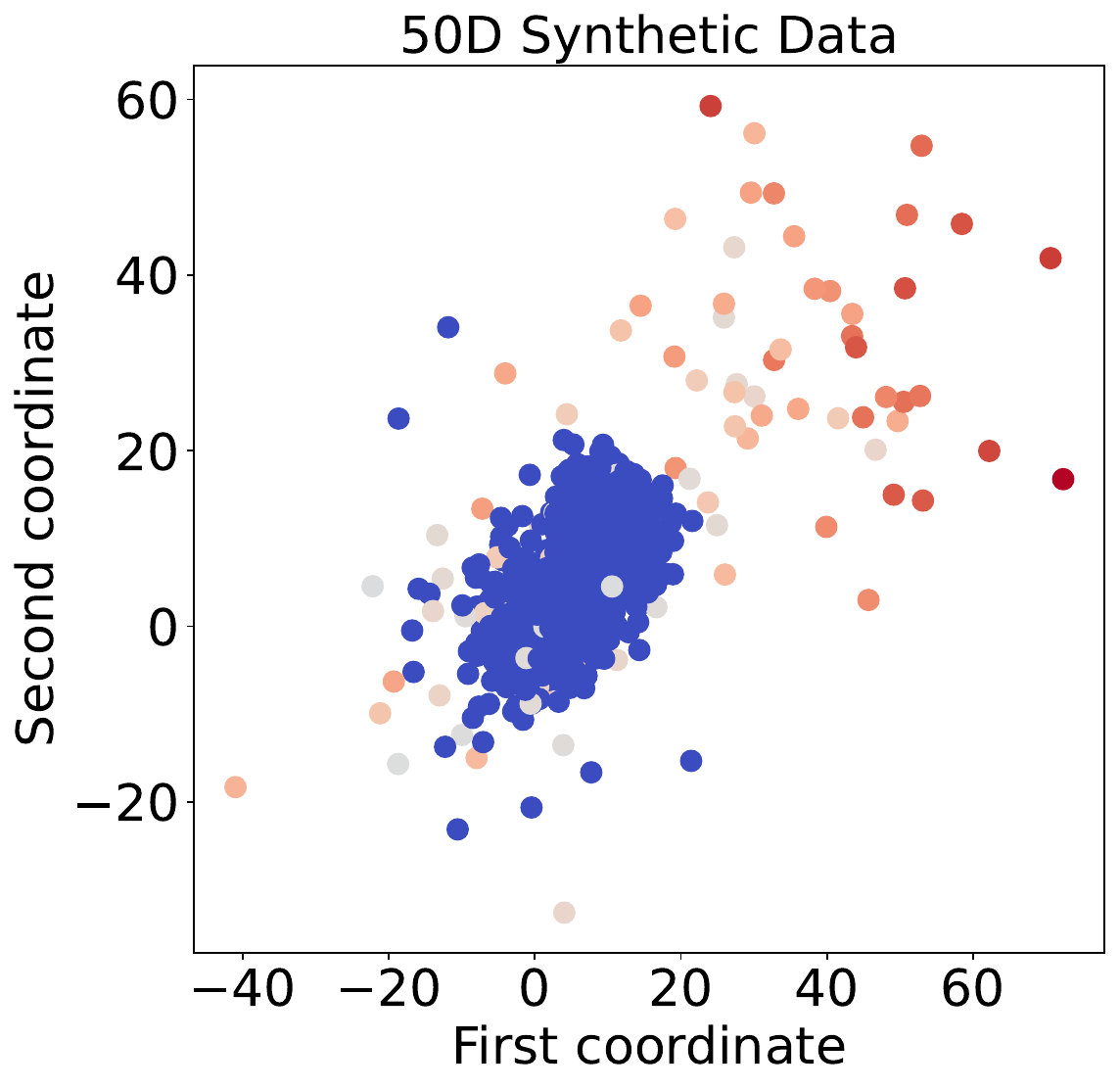} \label{fig:synthetic-data-2-10}}
	\subfloat[h][After 20 SGD steps]{\includegraphics[scale=0.3]{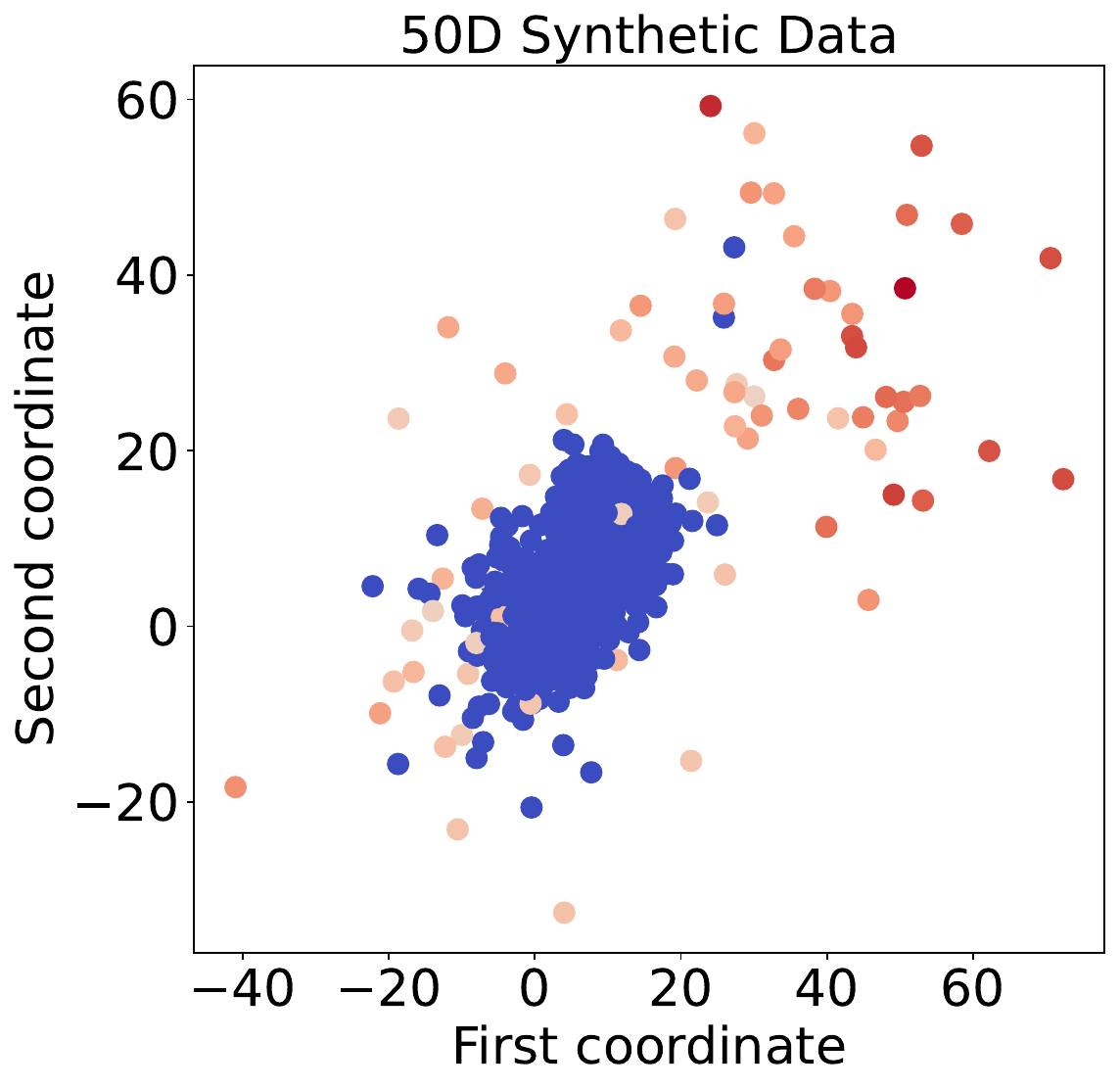} \label{fig:synthetic-data-2-50}}
	\caption{A 50-dimensional synthetic dataset. This dataset has 1,000 points generated from a mixture of two 50-dimensional Gaussian distributions (we only plot the first two features for visualization). The colour of each data point is from cold to warm indicating the corresponding value of the sparse eigenvector computed by our algorithm (\ref{alg:ads}). The sparsity parameter $K$ is set to 120 (the exact number of anomaly points). The epoch size $E$ is displayed under each sub figure from (a) to (h). The remaining parameters are set to their default values.} \label{fig:synthetic-data-2}
\end{figure*}

\subsection{Detecting Novelties in Synthetic Data}
\label{subsec:synthetic}
We first test the reaction of our algorithm to novel data points in both a low and a high dimensional setting in Figure \ref{fig:synthetic-data-1} and \ref{fig:synthetic-data-2} respectively. For the low-dimensional case we generate 100 uniformly distributed 2D points and make some of them anomalies by adding random noises to their coordinates. The high-dimensional dataset contains 1,000 points drawn from a mixture of two 50D Gaussian distributions, and the anomalies are synthesized in the same way as for the 2D case. It is interesting to see that our algorithm can locate a majority of deviating data points in both datasets after just one gradient step (\ref{eqn:component-2}). For the 2D dataset, the third step manages to find all ground-truth anomalies while leaving one false positive, which is then corrected by the following step. Similar iterative corrections can be observed for the 50D dataset, with a default number of 20 steps successfully identifying over 95\% of true anomalies. The default parameters of our algorithm work well in both settings, leaving just the sparsity value $K$ to choose by a user.

\begin{figure}
	\centering
	\subfloat[a][Precision, Recall, F-Measure]{\includegraphics[scale=0.3]{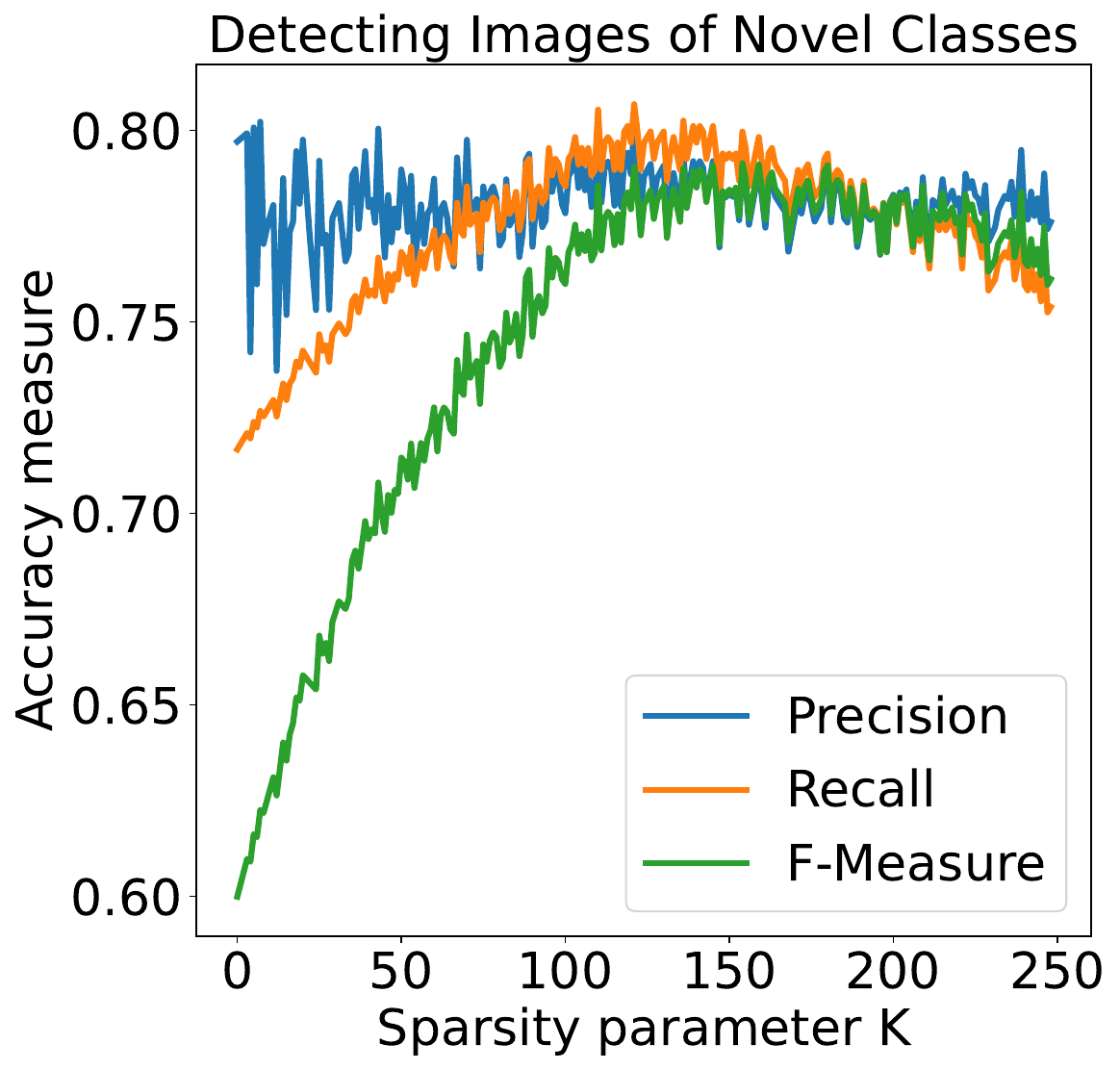} \label{fig:tiny-imagenet-train-prf}}
	\subfloat[b][Overlaid on t-SNE embeddings]{\includegraphics[scale=0.3]{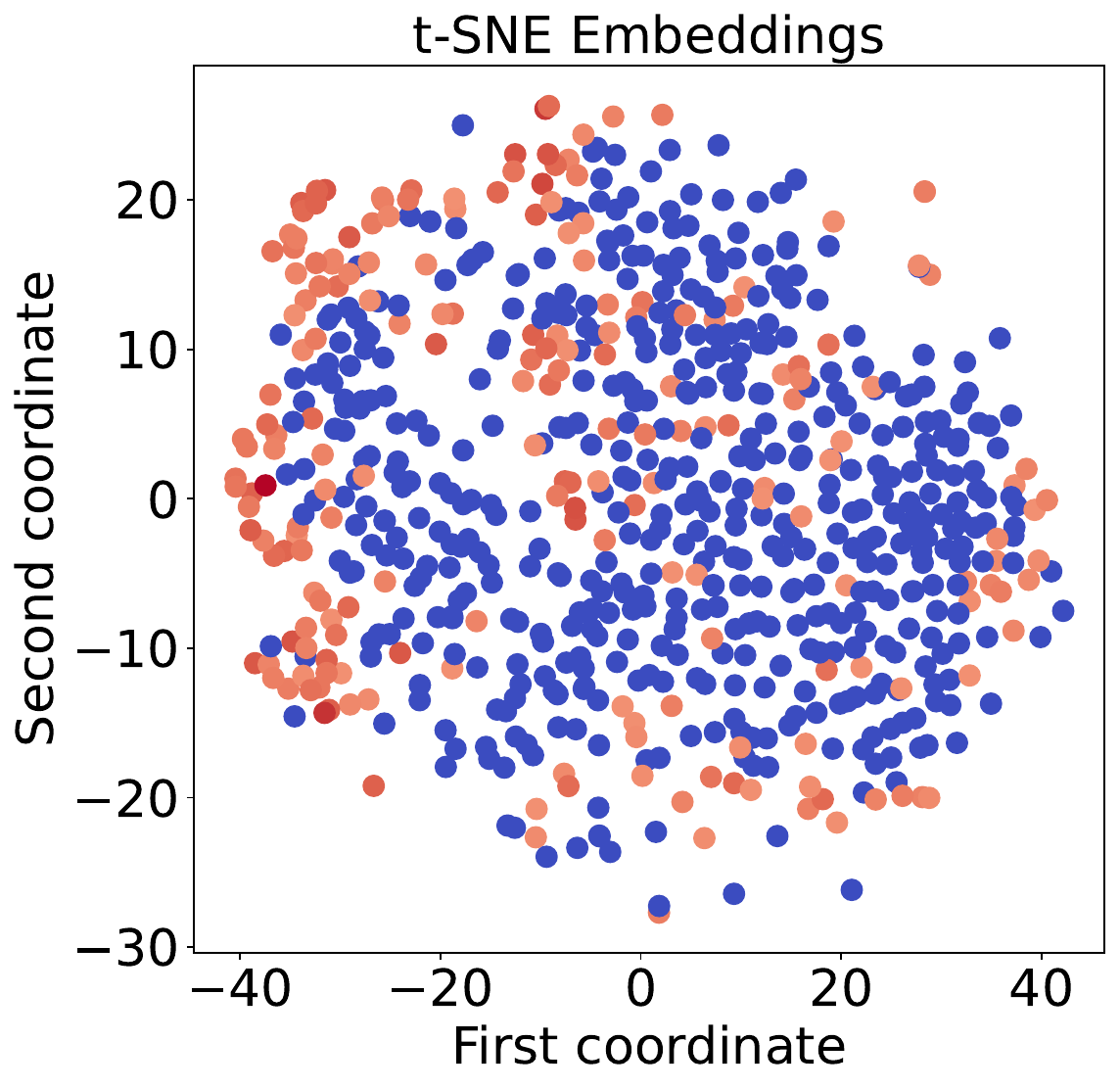} \label{fig:tiny-imagenet-train-tsne}}
	\caption{Tiny-ImageNet train set (500 images from the class Egyptian cat-n02124075 and 1 randomly chosen image from each of the remaining 119 classes) results. (a) shows the accuracy of our computed sparse eigenvector used for differentiating normal and novel images with respect to each value of $K$ - an image is classified to be novel if its computed weight is non-zero. (b) shows the weight of each image along with its t-SNE embeddings derived from the MobileNet-V2 extracted 1,000-dimensional features when $K$ is set to 119 as the ground-truth number of anomalies.} \label{fig:tiny-imagenet-train}
\end{figure}

\begin{figure}
	\centering
	\subfloat[a][Overlaid on PCA embeddings]{\includegraphics[scale=0.3]{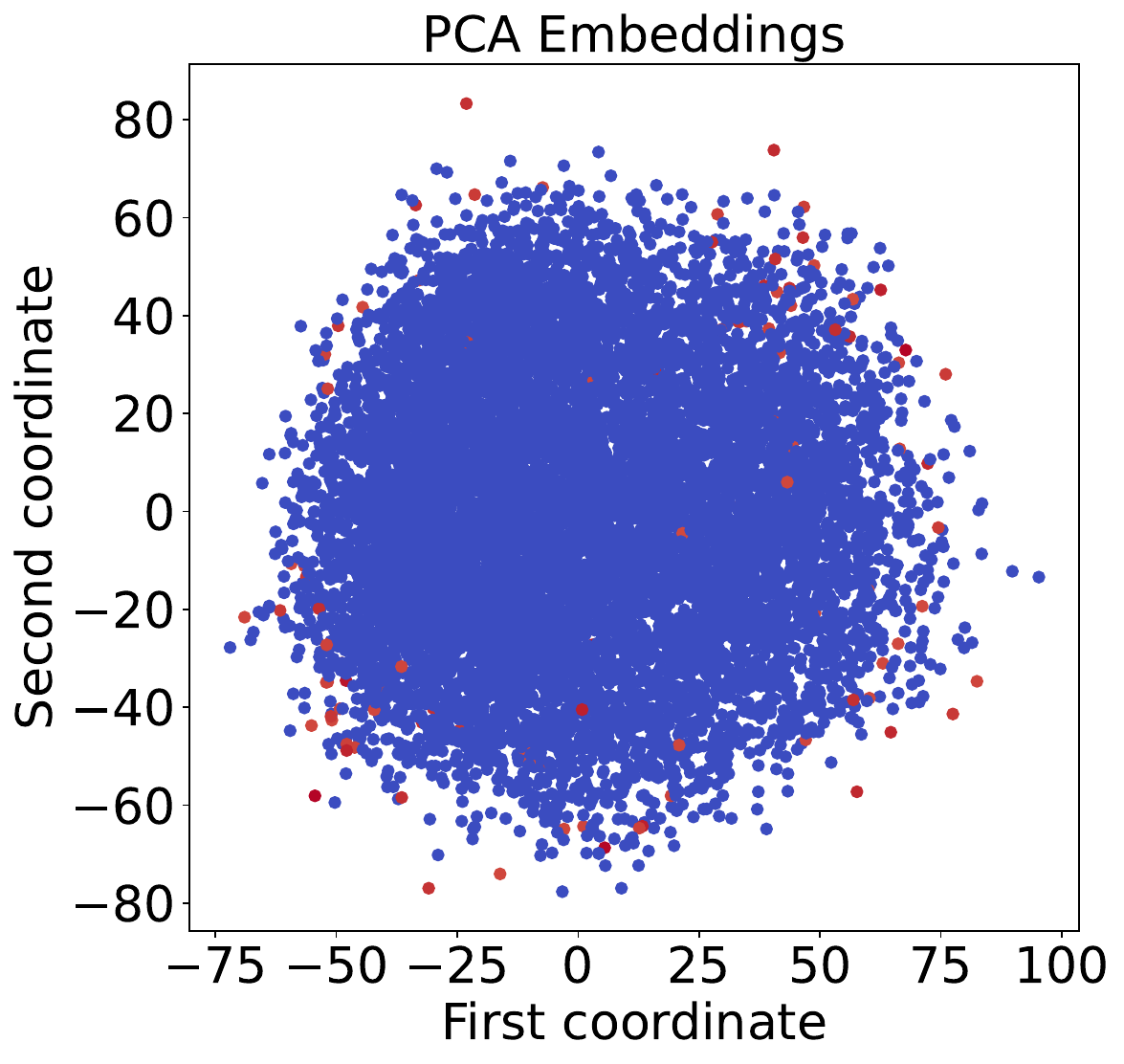} \label{fig:tiny-imagenet-test-pca}}
	\subfloat[b][Overlaid on t-SNE embeddings]{\includegraphics[scale=0.3]{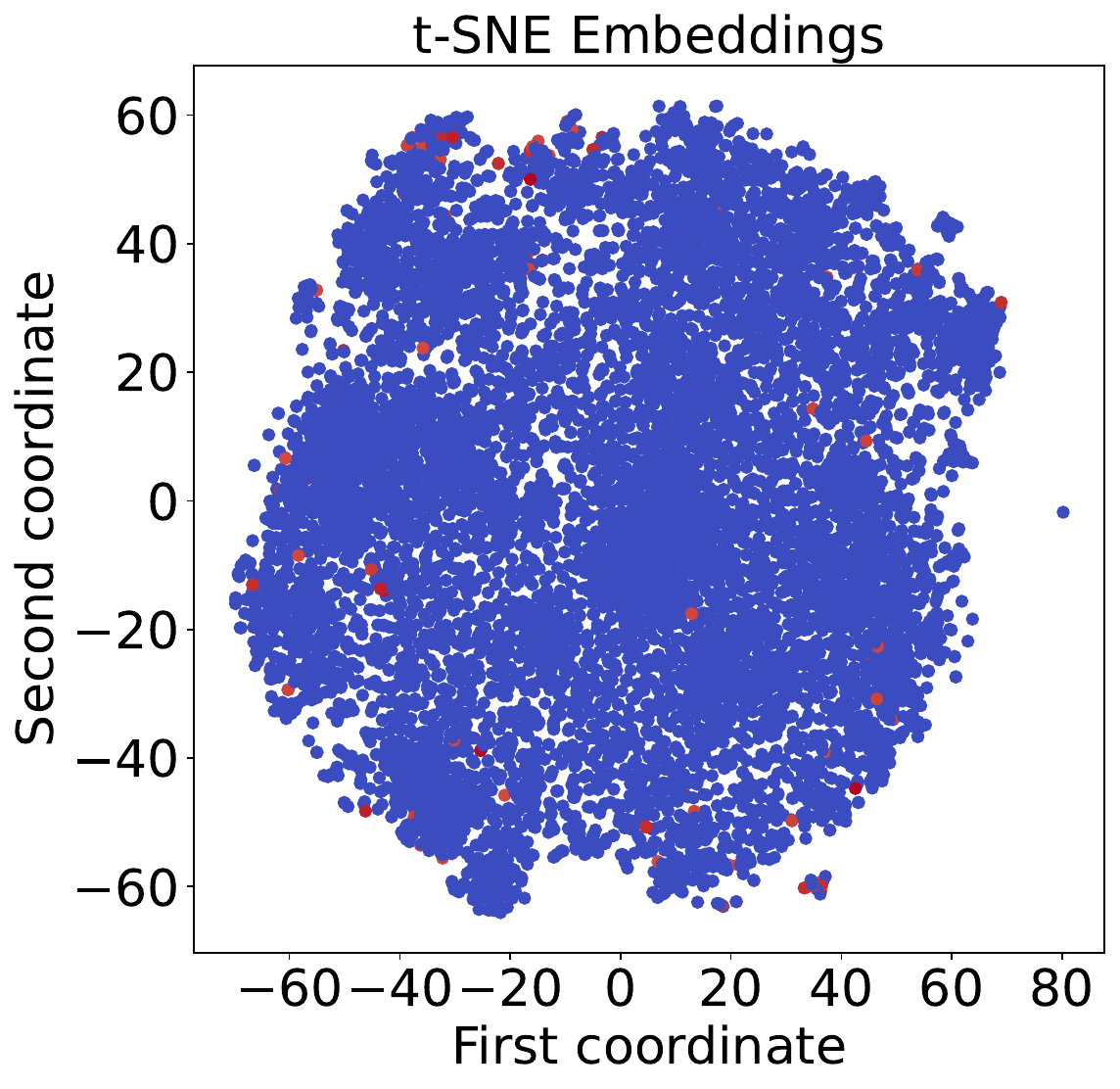} \label{fig:tiny-imagenet-test-tsne}}
	
	\subfloat[c][0.1063]{\includegraphics[scale=1.7]{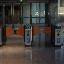} \label{fig:tiny-imagenet-test-sample1}}
	\subfloat[d][0.1061]{\includegraphics[scale=1.7]{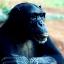} \label{fig:tiny-imagenet-test-sample2}}
	\subfloat[e][0.1057]{\includegraphics[scale=1.7]{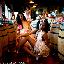} \label{fig:tiny-imagenet-test-sample3}}
	
	\subfloat[f][0.1056]{\includegraphics[scale=1.7]{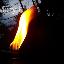} \label{fig:tiny-imagenet-test-sample4}}
	\subfloat[g][0.1045]{\includegraphics[scale=1.7]{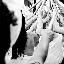} \label{fig:tiny-imagenet-test-sample5}}
	\subfloat[h][0.1044]{\includegraphics[scale=1.7]{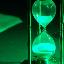} \label{fig:tiny-imagenet-test-sample6}}
	\caption{Tiny-ImageNet test set (10,000 images) results. (a) and (b) visualize our computed novelty weight for each image along with its PCA and t-SNE embeddings derived from the MobileNet-V2 extracted 1,000-dimensional features. (c)-(h) show the 6 images with the highest weight (displayed under each corresponding sub figure) found by our algorithm with the sparsity $K=100$.} \label{fig:tiny-imagenet-test}
\end{figure}

\subsection{Identifying Novel Images from Real Datasets}
\label{subsec:novel-class}
We now apply our algorithm to the highly non-trivial Tiny-ImageNet dataset \cite{le2015tiny} and show the results in Figure \ref{fig:tiny-imagenet-train} and \ref{fig:tiny-imagenet-test}. We use the popular lightweight MobileNet-V2 \cite{sandler2018mobilenetv2} pre-trained neural network to extract a 1,000-dimensional feature vector for each image. The first experiment we do is mixing the 500 training images of one particular category with 1 randomly drawn image from each of the remaining 119 classes that serve as true novelties. By varying the sparsity $K$ from 1 to 250 while keeping the other parameters as default, we can run our algorithm to produce a series of anomaly predictions and measure their accuracy using the class-balanced precision, recall, and f-measure metrics. Figure \ref{fig:tiny-imagenet-train-prf} shows that the precision of our algorithm for novel class identification remains high under a varying number of images found. The recall keeps increasing when $K$ is increased from 1 to around 150 and then drops slightly after that, which shows that our algorithm works well when the sparsity is set just a bit higher than the actual number of novel images in the dataset. The f-measure curve also reflects the overall monotonic performance increase of our algorithm when $K$ is not set too high. The local fluctuations of these metrics are due to the non-convexity of the problem (\ref{eqn:relaxation}) and the inherent randomness of SGD \ref{eqn:component-2}, which can be simply smoothed out by averaging multiple runs with different seeds. Figure \ref{fig:tiny-imagenet-train-tsne} visualizes our computed sparse eigenvector along with a t-SNE embedding \cite{van2008visualizing} of the dataset. It is interesting to note that the identified image anomalies concentrate around the periphery as well as low-density regions of the empirical data distribution in the feature space. To reproduce this on a held-out test set, we also run our algorithm on the 10,000 Tiny-ImageNet test images (with no labels available) with $K=100$ and show the results in Figure \ref{fig:tiny-imagenet-test}. While principal component analysis (PCA) does not seem able to delineate the dataset structures as t-SNE do, both visualizations show that our algorithm can accurately identify images of low probability from the given collection, with some examples displayed in Figure \ref{fig:tiny-imagenet-test-sample1} to \ref{fig:tiny-imagenet-test-sample6}.

\section{Conclusion}
\label{sec:conclusion}
We have proposed a simple yet effective algorithm for finding images of distinct characteristics within a large collection. The key of our algorithm is formulating the problem as solving for the $K$-densest subgraph within a perceptual distance-weighted complete graph of images. While the original problem is NP-hard, we have successfully relaxed it to be a sparse eigenvector problem, which we have effectively solved using stochastic gradient descent with sparsity clipping. Unlike any supervised learning methods, our algorithm is training-free and therefore does not require human-annotated labels to work. It also scales well with the number of the images in a collection in terms of both space and time complexity. The experimental results show the accuracy of our algorithm for picking up novel images.

An important future work would be establishing a theoretical analysis of the convergence properties of our algorithm (\ref{alg:ads}). The SGD method has previously been applied to PCA problems \cite{oja1985stochastic} with certain convergence guarantee of eigen-decomposition \cite{shamir2016convergence}. However, the analysis cannot extend to our case as we work with distance rather than covariance matrices.

%\section*{Appendix - Python Implementation}
%\label{sec:appendix}
%\lstset{language=Python}
%\lstset{frame=lines}
%\lstset{caption={Implementation of our algorithm (\ref{alg:ads}) in Python}}
%\lstset{label={lst:code_direct}}
%\lstset{basicstyle=\footnotesize}
%\lstinputlisting[language=Python]{ads_paper.py}

%\inputminted[fontsize=\footnotesize]{python}{ads_for_paper.py}





\bibliographystyle{elsarticle-num}
\bibliography{references}

%\section*{Biographies}
%\textbf{Shanfeng Hu} is a Senior Lecturer at the Computer and Information Sciences Department, Northumbria University. He graduated as a PhD from the department in March 2020 and subsequently joined it as an academic staff in June the same year. His research interest includes machine learning, decision-making, and causal inference.

\end{document}